# A Review of Deep Learning Techniques for Markerless Human Motion on Synthetic Datasets

Doan Duy Vo, Russell Butler

Department of Computer Science, Bishop's University

Sherbrooke, Quebec, Canada

Emails: dvo20@ubishops.ca, rbutler@ubishops.ca

25 December 2021

**Abstract**

*Markerless motion capture has become an active field of research in computer vision in recent years. Its extensive applications are known in a great variety of fields, including computer animation, human motion analysis, biomedical research, virtual reality, and sports science. Estimating human posture has recently gained increasing attention in the computer vision community, but due to the depth of uncertainty and the lack of the synthetic datasets, it is a challenging task. Various approaches have recently been proposed to solve this problem, many of which are based on deep learning. They are primarily focused on improving the performance of existing benchmarks with significant advances, especially 2D images.*

*Based on powerful deep learning techniques and recently collected real-world datasets, we explored a model that can predict the skeleton of an animation based solely on 2D images. Frames generated from different real-world datasets with synthesized poses using different body shapes from simple to complex. The implementation process uses DeepLabCut on its own dataset to perform many necessary steps, then use the input frames to train the model. The output is an animated skeleton for human movement. The composite dataset and other results are the "ground truth" of the deep model.*

## 1. Introduction

Behavioral quantification is important for many applications using computer vision. Imaging is perhaps the most common and widely used method because it allows non-invasive, high-resolution behavioral observations in a variety of settings [1,2,3]. However, extracting specific aspects of a behavior from video and further analyzing them always pose a challenging computational problem. Therefore, finding a robust and accurate way to measure human behavior has always been the focus of researchers. Recent advances in deep learning have shown that the process of quantifying behavior has been greatly simplified [4]. One of the great advantages of deep learning-based methods is that they are extremely flexible and allow researchers to define what to track [5].

Behavioral quantification is closely related to the process of identifying and tracking the 3D locations of joints in the human body. Human motion capture involves recording the movement of a moving human body and converting that movement into an abstract digital format. First, we used a single webcam with only one character to collect the real-world dataset. The datasets include the easy pose as T-pose, A-pose, standing and seating; the inter pose, mostly walking and running and the hard pose includes all the postures with high complexity like yoga, push-ups, and activities. Record difficulty progression is especially useful for curriculum learning applications [7]. By using DeepLabCut, an open-source toolbox for providing behavior tracking and markerless pose estimation [6], is used to generate a composite dataset and labeled 2D images. The animated skeleton for each frame is the result of performing deep learning built into the DeepLabCut toolbox. This process is carried out for the purpose of predicting the articulated joint locations of a human body from an image or a series of images from synthetic datasets. The result is a video in .mp4 format with labels created for visualization purposes with a network predictive label showing the trajectory of all body parts throughout the video.

The overall goal of our research is to use DeepLabCut toolbox to predict the joint rotation and position parameters to create an animated skeleton based solely on video camera data, without the use of special markers or tracking equipment. Therefore, the main structural paper will introduce the process of generating real video to 2D image frames in section 2. Section 3 will discuss a deep learning model to emphasize the importance of research on this topic and use them to pre-training input frames. Then, section 4 discusses the main challenges and the results obtained during the experiment. Finally, section 5 summarizes the work presented in this paper.

## 2. Data Collection

### 2.1. Synthetic Dataset

Ground Truth movements were used to record human behavior in real-world datasets. The recorded video contains human movements with different actions. All data is collected indoors, and the sequence does not contain external occlusions or significant disruption, but some of the challenges we face tend to disrupt background subtraction. The human performs four different performances, repeated many times: idle, walking, acting, and freestyle. As a result, three datasets were collected, including the easy pose as T-pose, A-pose, standing and seating; the inter pose, mostly walking and running and the hard pose includes all the postures with high complexity and varied set of motions like yoga, push-ups, and activities. They are performed under realistic indoor conditions that can be applied to most of the proposed pose and motion estimation techniques.

By using DeepLapCut, a pipeline designed to generate patterns from a 2D composite image of the human body shapes using different poses direct video without calibrate or retarget animation. Through experimentation, we have discovered that the key to a successful feature detector is the selection of different frames that are typical of human behavior and need to be labeled. The tool also allows editing the number of frames to extract for each input video. Label the extracted frames performed during creating the training dataset.

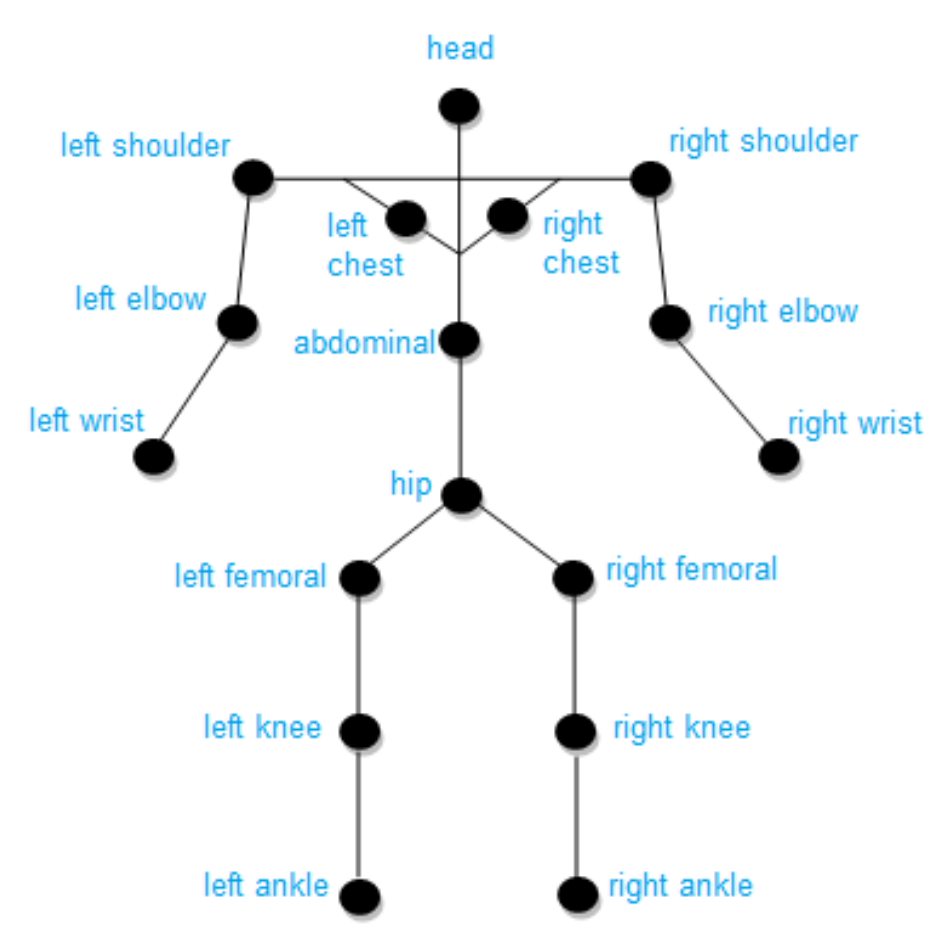

*Figure 1. Distribution diagram of label joint node*

The human skeleton is divided into 17 critical joints, and the whole-body motion posture is built through integration and calculation of the motion data of 17 joints. Therefore, 17 labels are pre-made on body parts including ankle, knee, femoral, wrist, elbow, shoulder, chest for each left and right side respectively, hip, abdominal and head. Figure 1 is a simulation of a hierarchical model of joints in the human body where black dots represent joints as labeled positions. Based on this model, a labeling process is performed (see the example in Figure 2) and the dataset is saved in the correct format for future steps. Following these steps, a synthetic dataset was created to store the 2D image labeled as the animated skeleton for each frame.

*(a) 2D image labeled of easy pose dataset*

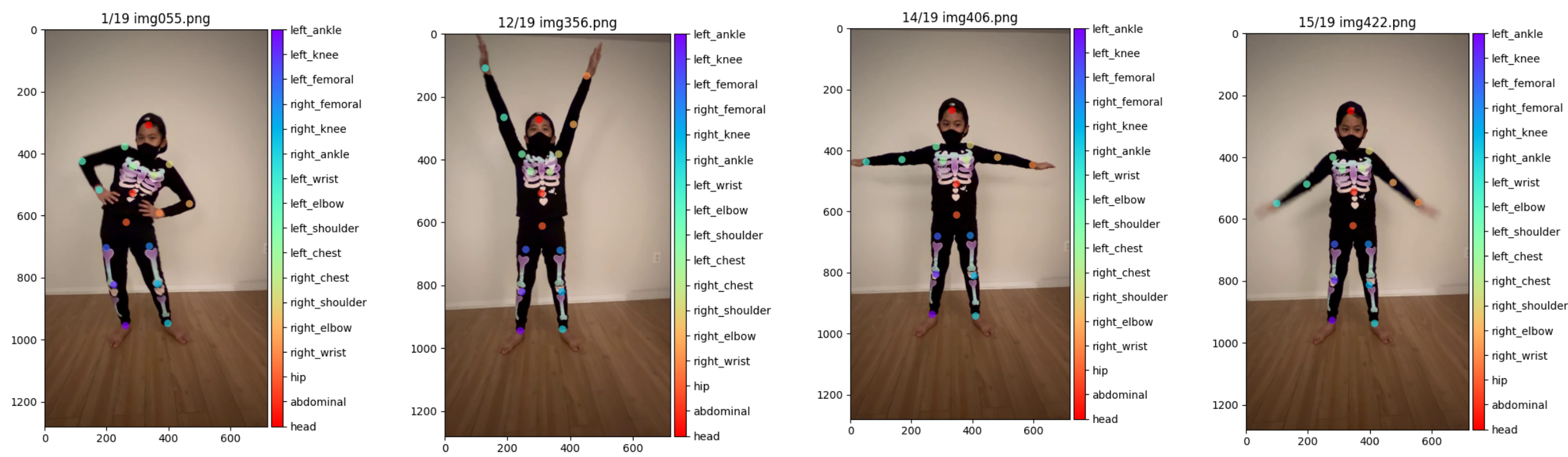

*(b) 2D image labeled of inter pose dataset*

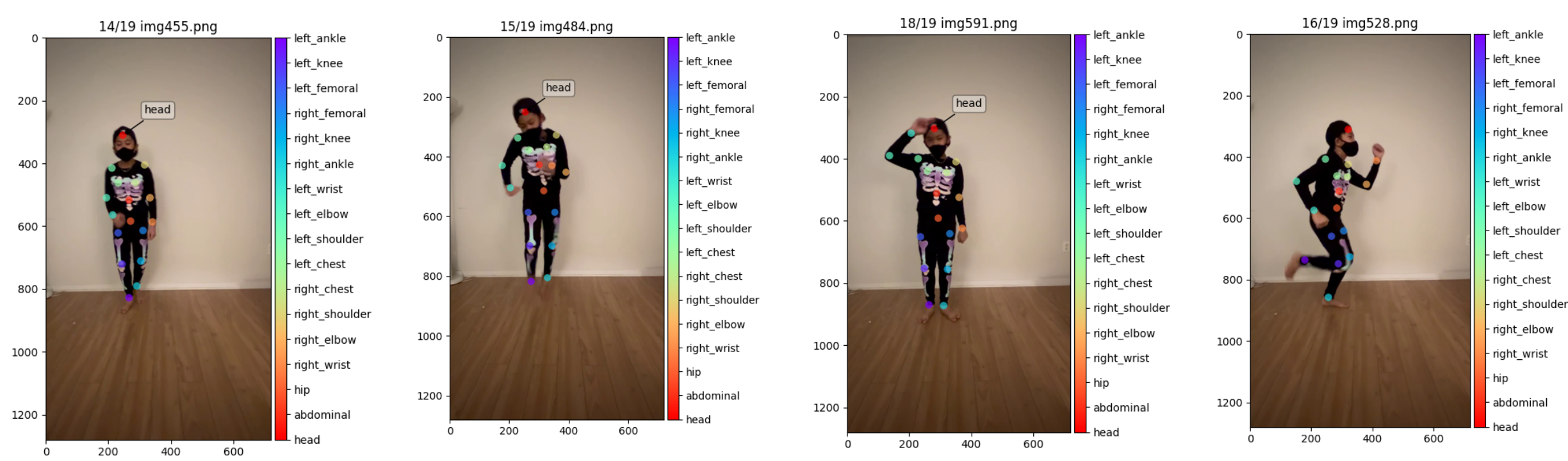

*(c) 2D image labeled of hard pose dataset*

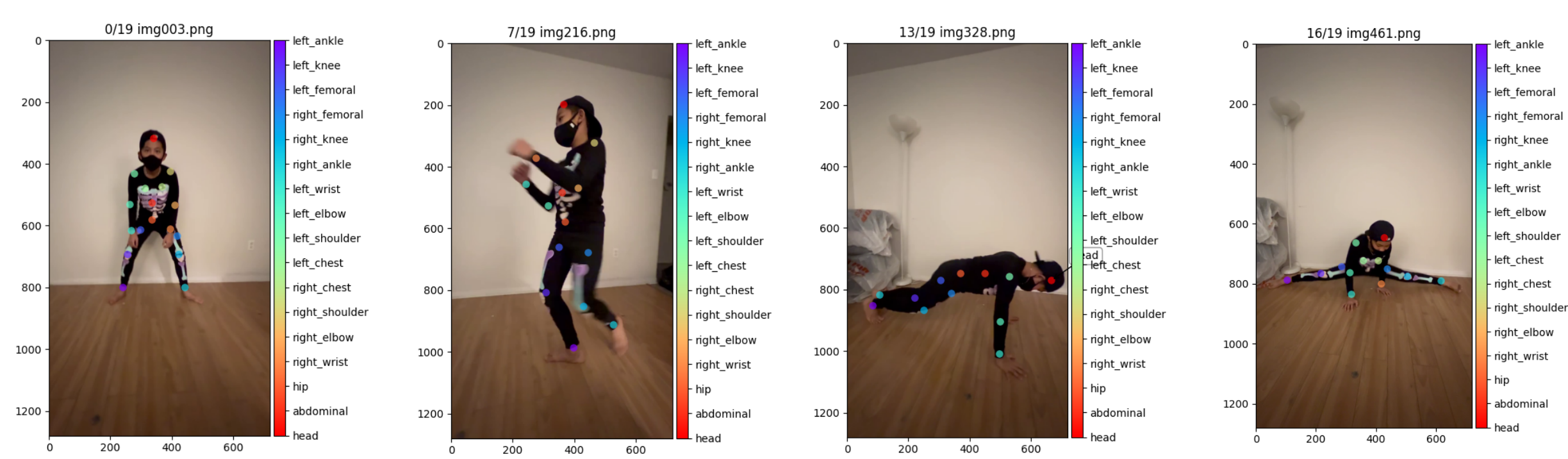

*Figure 2. 2D images labeled*

Labelling a frame is a step that requires accuracy and synchronization, as it is very important to label correctly in the same place. If some positions in the frame are hidden, it may be acceptable to skip the labeling procedure at that position. Therefore, choosing the most appropriate and well-stored frame for behavior is useful for training, as identification is one of the most important parts of creating a training dataset.

| scorer | resnet50_shuf | resnet50_shuf | resnet50_shuf | resnet50_shuf | resnet50_shuf | resnet50_shuf | resnet50_shuf | resnet50_shuf | resnet50_shuf | resnet50_shuf | resnet50_shuf | resnet50_shuf | resnet50_shuf | resnet50_shuf | resnet50_shuf |
|---|---|---|---|---|---|---|---|---|---|---|---|---|---|---|---|
| bodyparts | ankle1 | ankle1 | ankle1 | knee1 | knee1 | knee1 | hip1 | hip1 | hip1 | hip2 | hip2 | hip2 | knee2 | knee2 | knee2 |
| coords | x | y | likelihood | x | y | likelihood | x | y | likelihood | x | y | likelihood | x | y | likelihood |
| labeled-data\IMG_0096\img231.png | 332.5950012 | 346.4248352 | 0.061548203 | 383.1376038 | 477.3145142 | 0.044341087 | 16.36784172 | -0.229787827 | 0.035320073 | 10.23653412 | 2.192389727 | 0.038764417 | -7.998227119 | 2.531293869 | 0.074347079 |
| labeled-data\IMG_0096\img335.png | 366.2595215 | 441.8001099 | 0.070232362 | 361.9819641 | 443.7246399 | 0.055924267 | 362.8136597 | 443.3826904 | 0.047323197 | 363.552002 | 444.8964233 | 0.0495857 | -8.067686081 | 2.462677956 | 0.073152214 |
| labeled-data\IMG_0096\img063.png | 331.0294189 | 361.094696 | 0.073258162 | 10.95885658 | 1276.048706 | 0.054527789 | 330.6713867 | 332.2281799 | 0.044273138 | 334.5252686 | 332.1541138 | 0.047741532 | -8.150652885 | 2.05121994 | 0.079113454 |
| labeled-data\IMG_0096\img265.png | 317.4851074 | 474.1382446 | 0.066217363 | 12.09158993 | 1275.988403 | 0.050743252 | 315.9909363 | 467.0641479 | 0.045904487 | 329.433136 | 348.9205933 | 0.043857604 | -7.691561699 | 1.956284046 | 0.072827041 |
| labeled-data\IMG_0096\img132.png | 350.4554749 | 363.6315002 | 0.058001578 | 10.85644531 | 1275.836792 | 0.055955827 | 395.3374634 | 378.6596069 | 0.045919687 | 395.3800964 | 380.3094482 | 0.04217279 | -8.175878525 | 2.326956272 | 0.073097855 |
| labeled-data\IMG_0096\img351.png | 380.6958313 | 460.4442444 | 0.055573881 | 9.801513672 | 1275.944336 | 0.047910541 | 16.48183441 | -0.496304989 | 0.039571375 | 282.1497192 | 878.5777588 | 0.0416044 | -8.277057648 | 2.184408665 | 0.077753514 |
| labeled-data\IMG_0096\img152.png | 398.8226318 | 377.9899597 | 0.065701962 | 396.4528809 | 380.1190491 | 0.04374823 | 394.3954163 | 378.0456238 | 0.04392153 | 395.7487488 | 380.5853882 | 0.048846453 | -8.070039749 | 2.138823271 | 0.072890371 |
| labeled-data\IMG_0096\img101.png | 331.7119446 | 362.5954285 | 0.090387613 | 11.559412 | 411.0495605 | 0.057288051 | 316.5441895 | 380.0693054 | 0.052182823 | 398.6968079 | 411.5429688 | 0.052600771 | -7.891150475 | 2.310299873 | 0.075922638 |
| labeled-data\IMG_0096\img358.png | 381.2197876 | 460.4522095 | 0.059171766 | 382.2346497 | 460.9538879 | 0.056177497 | 377.650177 | 456.519928 | 0.039086282 | 345.1730957 | 220.9837646 | 0.043748796 | 378.7697449 | 472.5927429 | 0.077349216 |
| labeled-data\IMG_0096\img047.png | 332.2039185 | 347.133728 | 0.076179504 | 10.95476341 | 1276.279053 | 0.06028387 | 330.9757385 | 346.7367859 | 0.043084025 | 413.4247437 | 393.2244568 | 0.045767665 | -8.01511097 | 2.235839367 | 0.076878637 |
| labeled-data\IMG_0096\img299.png | 317.5386963 | 474.3696594 | 0.06782639 | 12.13589382 | 1275.666748 | 0.051767796 | 315.1959839 | 475.0462952 | 0.042621523 | 10.39332581 | 1.940620422 | 0.040031165 | -8.053044319 | 2.353956938 | 0.077572286 |
| labeled-data\IMG_0096\img175.png | 332.613678 | 345.9727478 | 0.073651552 | 11.26898003 | 1276.654907 | 0.04041478 | 375.9110107 | 458.0720825 | 0.040296912 | 382.1602478 | 444.3212585 | 0.040089786 | -8.297294617 | 2.240172148 | 0.073851466 |
| labeled-data\IMG_0096\img117.png | 331.4884338 | 347.1121216 | 0.058401138 | 9.852176666 | 459.7755432 | 0.058370352 | 316.6655884 | 380.8707275 | 0.042195976 | 415.1938171 | 408.7536926 | 0.043538094 | -8.360219955 | 2.358124018 | 0.083210915 |
| labeled-data\IMG_0096\img251.png | 333.4929504 | 363.2973938 | 0.064364552 | 11.76252747 | 1275.471924 | 0.052084923 | 283.1618347 | 443.040802 | 0.039167166 | 286.6871948 | 425.3531799 | 0.040564865 | -7.901388168 | 2.091419935 | 0.073589414 |
| labeled-data\IMG_0096\img083.png | 332.116333 | 362.9113464 | 0.081968099 | 10.59095001 | 1275.901123 | 0.064520806 | 317.1332092 | 379.3300171 | 0.041642904 | 398.37854 | 411.6311035 | 0.049499631 | -8.252846718 | 2.000121593 | 0.082396626 |
| labeled-data\IMG_0096\img128.png | 350.4564514 | 363.9376831 | 0.056788772 | 11.20575714 | 1276.005859 | 0.059582889 | 395.2371521 | 378.3322449 | 0.04011482 | 10.20850182 | 2.051493645 | 0.039695024 | -8.106217384 | 2.354981422 | 0.076713204 |
| labeled-data\IMG_0096\img038.png | 331.9135437 | 346.6182556 | 0.072025746 | 10.63271332 | 1276.681274 | 0.05807963 | 330.7931824 | 346.8624268 | 0.037634671 | 10.31821823 | 2.118758917 | 0.039394855 | -8.119765282 | 2.290288448 | 0.073303491 |
| labeled-data\IMG_0096\img261.png | 366.7522888 | 554.5628662 | 0.056118041 | 12.50396538 | 1276.64563 | 0.050589025 | 327.8453369 | 319.1828613 | 0.0476183 | 344.8921204 | 317.1112061 | 0.060589164 | -7.795343399 | 2.056432962 | 0.07329458 |
| labeled-data\IMG_0096\img002.png | 332.4949951 | 346.1307068 | 0.067525089 | 11.2051506 | 1275.907837 | 0.04518792 | 393.9529724 | 252.8136597 | 0.038944483 | 398.73526 | 411.2565613 | 0.046728373 | -8.353927612 | 2.260275364 | 0.073972464 |
| labeled-data\IMG_0096\img330.png | 300.7605286 | 443.0822449 | 0.073837221 | 362.7224731 | 476.5580444 | 0.044662267 | 362.260376 | 475.793335 | 0.043180019 | 363.4742126 | 477.2861023 | 0.04132691 | -8.351465225 | 2.387490988 | 0.07542637 |

*Table 1. Synthetic dataset with the animation skeleton for each frame*

## 2.2. Training Dataset

A good training dataset should consist of a sufficient number of frames that capture the breadth of the behavior and reflect different behaviors with respect to postures. Assessing the trained network typically requires thousands of training data, and researchers can repeat the training and testing process over and over to find the optimal training data set before making accurate predictions. Therefore, training datasets are usually labelled with highest possible accuracy and synchronization requirements. In other words, to build a robust model, the training dataset should include examples with a variety of human poses. Due to the challenges of traditional data collection approaches, including capturing human movements with cameras, data synthesis plays an important role as a generative model for synthesizing high quality depth frames with fully labeled ground truth frames. Therefore, several training experiments with varying numbers of training images to determine the effective size of the training set based on a collection of three synthetic datasets containing training samples evenly distributed across all postures.

To create training datasets, DeepLabCut provided the function to combine the labeled datasets and split them to create train and test datasets [6]. The training data is used to train the network, while the test data is used to evaluate the network. The model is initially fitted to the training data set and produces a result for each input frame of the training dataset, which is compared to the target. The parameters of the model are adjusted based on the result of the comparison and the specific training algorithm used. Finally, the test dataset is the dataset used to monitor the performance of the validation test dataset while providing an evaluation of the final model that fits the training dataset.

## 3. Deep Posture Analysis for Markerless Human Motion

### 3.1. Pre-trained and Evaluate Models

In this study, DeepLabCut was found to provide a pre-trained human pose model at MooelZoo [6]. However, the process implemented a retraining and evaluation model to understand how DeepLabCut works and use synthetic datasets for training. In practice, several factors matter such as the performance of the fine-tuned model of the task in question, the number of images that need to be characterized to fine-tune the network, and the rate of convergence of the optimization algorithm. Therefore, the pre-trained models can also be adapted to a particular application.

The process begins by merging all the extracted labeled frames and splitting them into subsets of test and train frames to create a training dataset. The pre-trained network is then end-to-end trained, adjusting the weights and using resnet_50 or 101 to predict the desired function. The evaluating the performance of the trained network then performed against the training and testing framework. The trained network can be used to analyze videos that produce the extracted pose file. If the trained network does not successfully generalize to hide data in the evaluation and analysis steps, then additional frames with poor results can be extracted and manually move the predicted labels to the ideal position.

Within the scope of this project, a markerless human motion capture system provides three datasets of synthetic depth data for training single pose estimation systems. Our three datasets contain varying complex shapes and poses with thousands of frames. Here, the size of the data is large and the data of all three datasets are highly similar. Therefore, this is an ideal situation for most effectively pre-training the model range. The best way to use a model is to maintain the model's architecture and the initial weights of the model. Then we can re-train this model using the weights as initialized in the pre-trained model. To achieve the highest possible accuracy during training, we stopped the initial training phase at min 2700 iterations. We then improve the resulting network per dataset with as much iteration as possible. In all the steps above, the test dataset is reserved for the final pose estimation task, so the evaluation result will be the validation set for each dataset. The evolution of our three-pose dataset is shown in Table 2.

| | Training iterations: | %Training dataset | Shuffle number | Train error(px) | Test error(px) | p-cutoff used | Train error with p-cutoff | Test error with p-cutoff |
|---|---|---|---|---|---|---|---|---|
| Easy pose dataset | 2700 | 95 | 1 | 82.28 | 35.69 | 0.01 | 82.28 | 35.69 |
| Inter pose dataset | 6200 | 95 | 1 | 34.13 | 31.29 | 0.01 | 34.13 | 31.29 |
| Hard pose dataset | 2850 | 95 | 1 | 211.14 | 263.53 | 0.01 | 211.14 | 263.53 |

*Table 2: Combined evaluation results*

The evaluation results for each shuffle in the training dataset are saved in the evaluation process. The results show that the distance between the marked and the predicted body parts is on the same human body. The human labels are plotted as a plus '+' and the prediction from DeepLabCut is plotted as 'o' with p-cutoff is 0.01, which is a confident prediction with likelihood > p-cutoff. Examples test and training plots from easy pose dataset are depicted in Figure 3. When benchmarking with different shuffles on the same training dataset, the shuffle index can adapt and iteratively evaluate the corresponding network process. If the generalization is inadequate, double-check the labeling process and repeat the training dataset again.

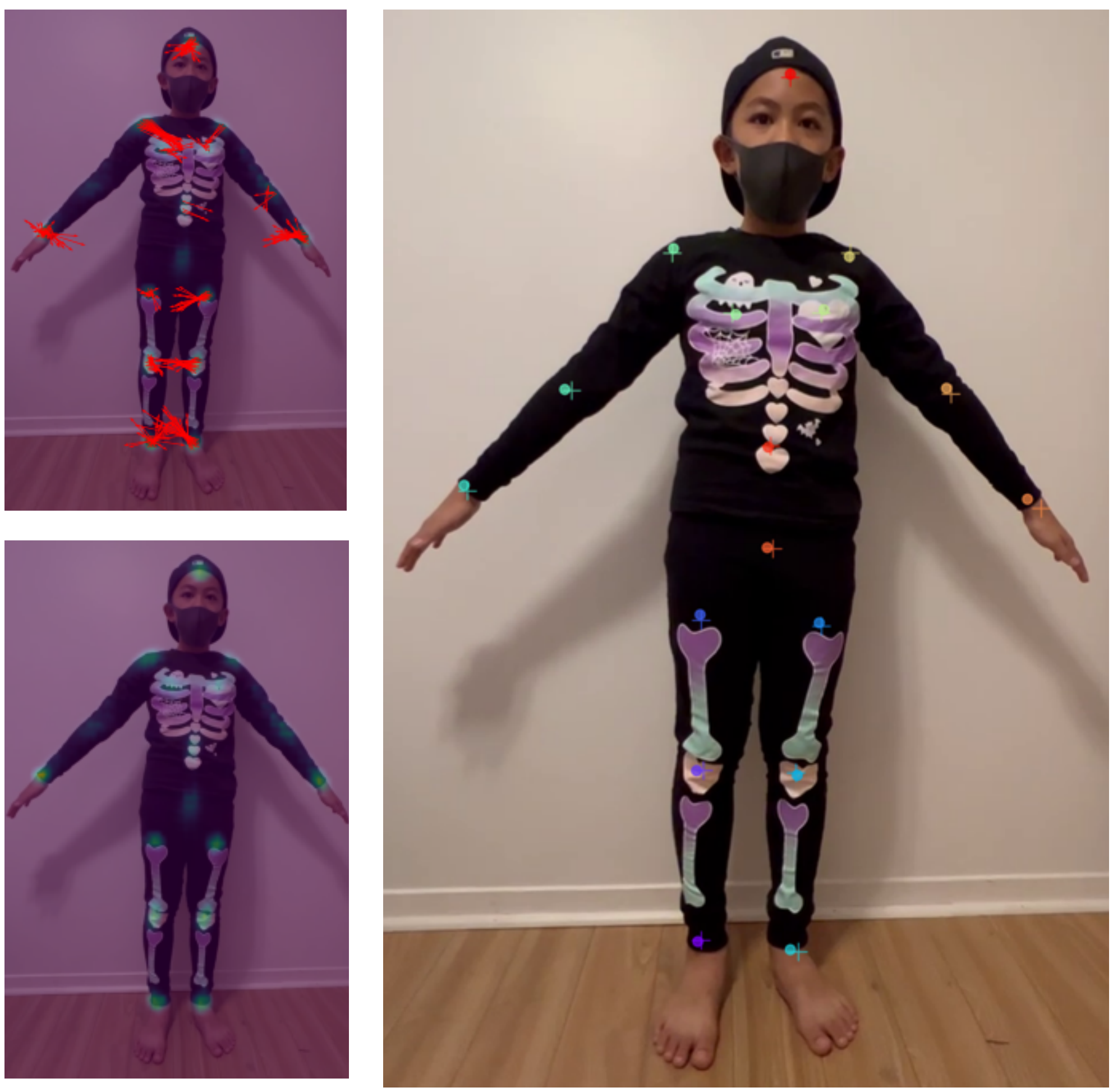

*Figure 3: Images from evaluation results*

### 3.2. Deep Posture Analysis and Plotting Results

The trained network can be used to analyze new videos. By selecting the best model from the evaluation results, the video analysis is performed based on the execution command. The result is an array [8] and if the save_as_csv flag is set to True, it will be saved in an efficient hierarchical data format (.hd5) and exported in a comma separated value format (.csv). These include the likelihood for each frame per body part, the name of the network, and body part name (x, y) label position in pixels. The labels for each body part across the video can also be plotted after the video has been analyzed. Trajectories of all the body parts throughout the entire video is plotted and each body part is identified by a unique color. Trajectories can also be easily imported into many programs for further behavioral analysis. They include trajectories filtered by body parts plotted across all frames in space; all body parts over time for each solid frame index are Y and dashed lines are X; body part likelihood across time over all frames; as well as histograms of consecutive coordinate differences with low values is the smallest jump between frames. The graph results are shown in Figure 4.

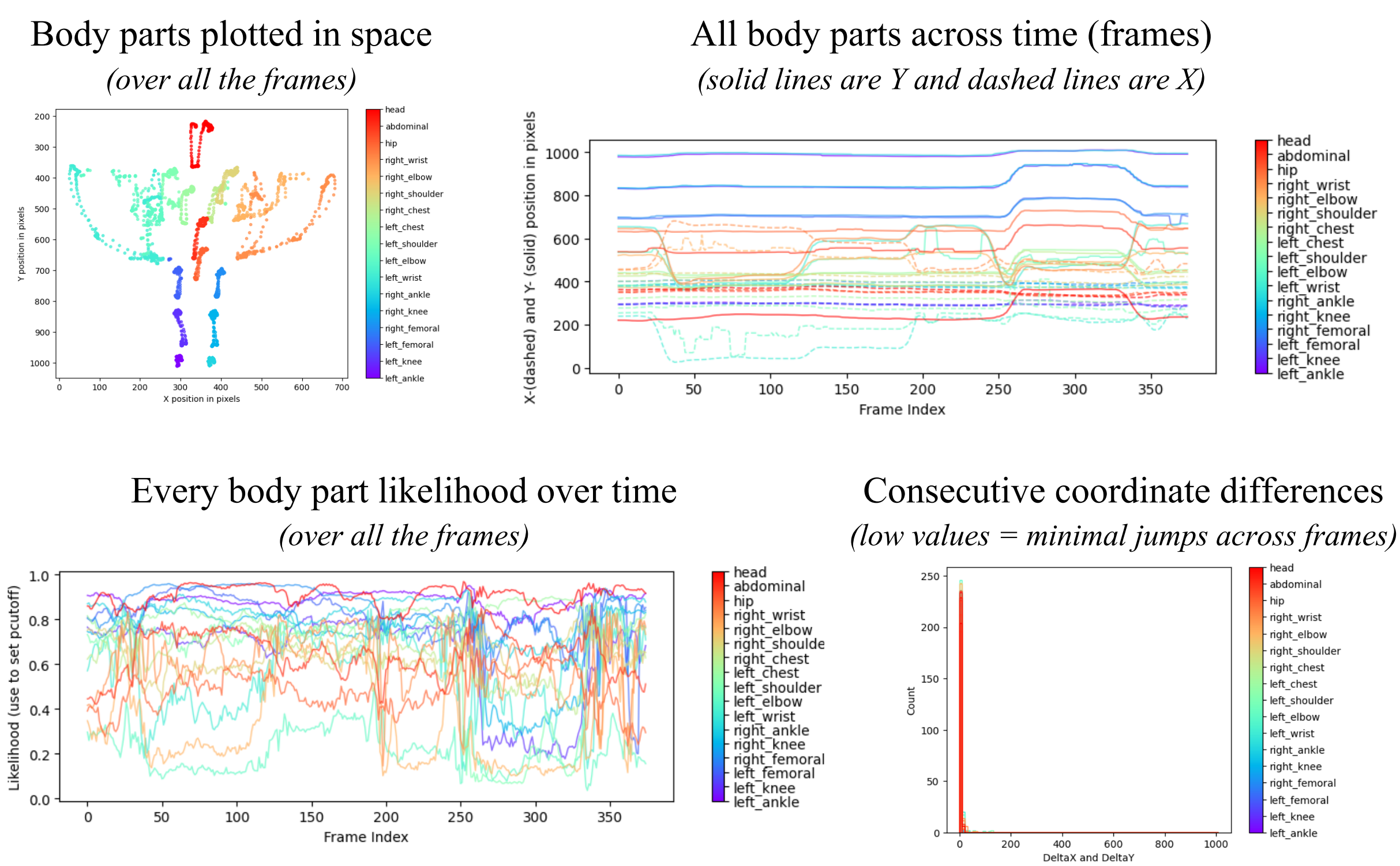

*Figure 4: The graphs plot the trajectories*

In addition, for visualization purposes, a labeled video will be created based on the extracted poses. It is saved as a video in .mp4 format with the labels predicted by the network. Depending on the configuration set in the config.yaml file, the video can add skeletons to connect points, and/or add a history of points for visualization or only the dots plotted. However, if save_frames = True is passed, the best quality videos will be created. Therefore, it is highly recommended to use save_frames = True when using trail points and draw_skeleton.

**4. Experiment Results**

The experimental process begins with collecting three real-world datasets. First, extract images from each video frame at a rate of 30fps at 200 frames. Reduce video from 30 fps to 15 fps on both the training and testing sets. Then select the 20 frames presented for the pose, define the centers of the 17 joints, and train the network using only the trajectories of those joints. Next, train and evaluate the performance of the proposed model. After training, we repeatedly refined the images and re-trained the network until we had frames that allowed good performance and generalization. Pre-train all three datasets to investigate the impact of various factors on the accuracy of the results. Finally, as a qualitative result, we provided the typical 3D poses predicted by labeled video. To run this whole process, use DeepLabCut with Python code to select training frames, check the human annotator labels, generate training data in the required format, as well as evaluate the performance on test frames. For 2D image frames extraction, this task can be performed by GUI or manually. The toolbox also contains code for extracting postures from the original videos using a trained feature detector to create labeled videos. Therefore, this toolbox is considered suitable for pre-trained a tailored network based on labeled images and can then perform automatic labeling for new data. Based on the above steps, we have created a version that meets the analytical needs of this project, based on three datasets. Check out the results here: https://github.com/DoanDuyVo/DeepLab_Human

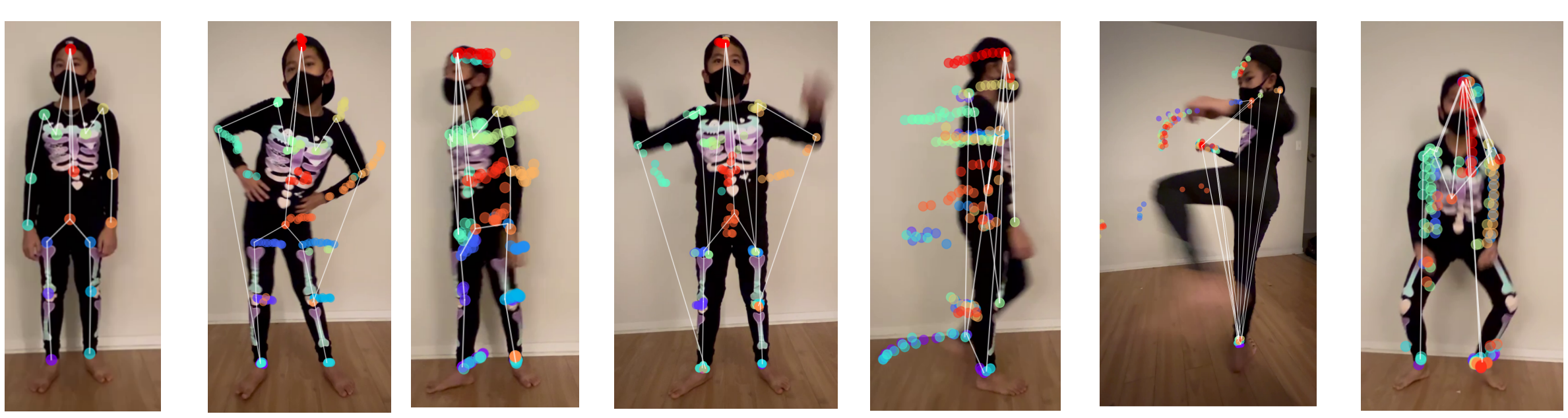

*Figure 5: Draw a skeleton connecting the nodes*

The original version of DeepLabCut can be found step-by-step in the user guide at https://github.com/DeepLabCut/DeepLabCut, which outlines the pipeline and workflow of a markerless pose estimation project. There are options to to perform markerless pose estimation on your local computer by both GUI and Python code, or in Colab which provide a pre-installed execution environment.

## 5. Conclusion

Recent advances in deep learning-based markerless pose estimation have been widely and quickly adopted. Part of this effect was facilitated by open-source code. By developing, sharing source code and pre-trained models in a public repository on GitHub, it's now free and easy to use on a large scale. These packages are progress-based and run code on a variety of platforms based on computer vision and AI technology with a strong open science culture. In addition, this wide range of applications of deep learning-based markerless pose estimation technology will contribute significantly to scientific research in areas such as medicine, virtual reality, animal behavior analysis, biomechanics, the game industries, robotics [9], etc. Therefore, research on deep learning-based markerless pose estimation will be a promising direction in the future.

As a result, this paper presented the process of using the synthetic datasets with varying shape and pose complexity in thousands of frames to evaluate and track human posture. These are the comprehensive datasets that include synced video from a single camera view, associated 3D ground truth, 2D pose frames, and 2D labeled images. This process is a prime example of capturing human movements, showing that training with synthetic data and manual data labeling with the model of DeepLabCut can lead to undifferentiated results. However, a deep convolutional neural network for motion tracking and pose estimation are more efficient with a state-of-the-art multiview pose estimation. All data and associated scripts used in this project are available to the research community. We hope that these datasets will lead to further advances in human joint motion estimation and provide an opportunity to determine the performance of current state-of-the-art algorithms.

In future work, we would like to extend our experiments to use a multi-view approach with multi cameras and expand the scope to include complex and continuous human movements such as ballet and martial arts. We are also interested in 3D reconstruction scenarios that reproduce 3D human movements on one side.

**Acknowledgments**

The authors would like to thank the DLC Team who created and maintained DeepLabCut as an open-source tool on GitHub with many benefits from suggestions and updates. We are grateful to Mathis Alexander and Brandon E. Jackson for suggestions on how to best use the TensorFlow implementation of DeeperCut and for showing us how to fix the errors we encountered during the execution of the experiment.